%% file: main.tex
\documentclass[runningheads]{llncs}

\usepackage[utf8]{inputenc} 
\usepackage[T1]{fontenc}    
\usepackage{hyperref}       
\usepackage{url}            
\usepackage{booktabs}       
\usepackage{amsfonts}       
\usepackage{nicefrac}       
\usepackage{microtype}      
\usepackage{color}
\usepackage[dvipsnames]{xcolor, colortbl}

\usepackage{graphicx}
\usepackage{amssymb}
\usepackage{amsmath}
\usepackage{pifont}
\usepackage{caption}
\usepackage{subcaption}
\usepackage{enumitem}
\usepackage{multirow}
\usepackage{wrapfig}

\newcommand*{\maroon}[1]{\textcolor{Maroon}{#1}}
\newcommand*{\teal}[1]{\textcolor{teal}{#1}}

\newcommand {\otoprule}{\midrule [\heavyrulewidth]}
\newcolumntype {+}{ >{\global\let\currentrowstyle\relax}}
\newcolumntype {^}{ >{\currentrowstyle }}
 \newcommand {\rowstyle}[1]{\gdef\currentrowstyle{#1} %
 #1\ignorespaces
 }
\newcommand{\tabhead}{\rowstyle{\bfseries}}

\usepackage{xspace}
\newcommand{\isic}{\text{\sc ISIC}\xspace}
\newcommand{\wb}{\text{\sc Waterbirds}\xspace}
\newcommand{\rn}{\text{ResNet18}\xspace}
\newcommand{\vit}{\text{ViT-B/16}\xspace}
\newcommand{\sscore}{\text{$\mathbf{s}$-score}\xspace}

\begin{document}
\title{An XAI-based Analysis of Shortcut Learning in Neural Networks}
%

\author{Phuong Quynh Le\inst{1}\orcidID{0000-0001-7658-0599} \and
Jörg Schlötterer\inst{1,2}\orcidID{0000-0002-3678-0390} \and
Christin Seifert\inst{1}\orcidID{0000-0002-6776-3868}}
\authorrunning{P. Q. Le , J. Schlötterer, and C. Seifert}
%
\institute{University of Marburg 
\and University of Mannheim 
\\
\email{\{phuong.le, joerg.schloetterer, christin.seifert\}@uni-marburg.de}}
\maketitle              
\begin{abstract}
Machine learning models tend to learn spurious features -- features that strongly correlate with target labels but are not causal. 
Existing approaches to mitigate models' dependence on spurious features work in some cases, but fail in others. In this paper, we systematically analyze how and where neural networks encode spurious correlations. We introduce the neuron spurious score, an XAI-based diagnostic measure to quantify a neuron's dependence on spurious features. We analyze both convolutional neural networks (CNNs) and vision transformers (ViTs) using architecture-specific methods. Our results show that spurious features are partially disentangled, but the degree of disentanglement varies across model architectures. Furthermore, we find that the assumptions behind existing mitigation methods are incomplete. Our results lay the groundwork for the development of novel methods to mitigate spurious correlations and make AI models safer to use in practice.

\end{abstract}

\input{./01_introduction}
\input{./02_background}
\input{./03_experimental-setup}

\input{./04_learning}
\input{./05-1_cnn-disentanglement}
\input{./05-2_cnn-entanglement}
\input{./05-3_cnn-modular}
\input{./06-1_vit-lastlayer}
\input{./06-2_vit-attention-heads}
\input{./07_discussion}

\input{./08_related-work}
\input{./09_conclusion}

\clearpage

%
%
\bibliography{export-data}
\bibliographystyle{splncs04}

\input{appendix}

\end{document}

%% file: 01_introduction.tex
\section{Introduction} \label{sec:intro}

\begin{figure}[t!hbp]
    \centering
    \includegraphics[width=\linewidth]{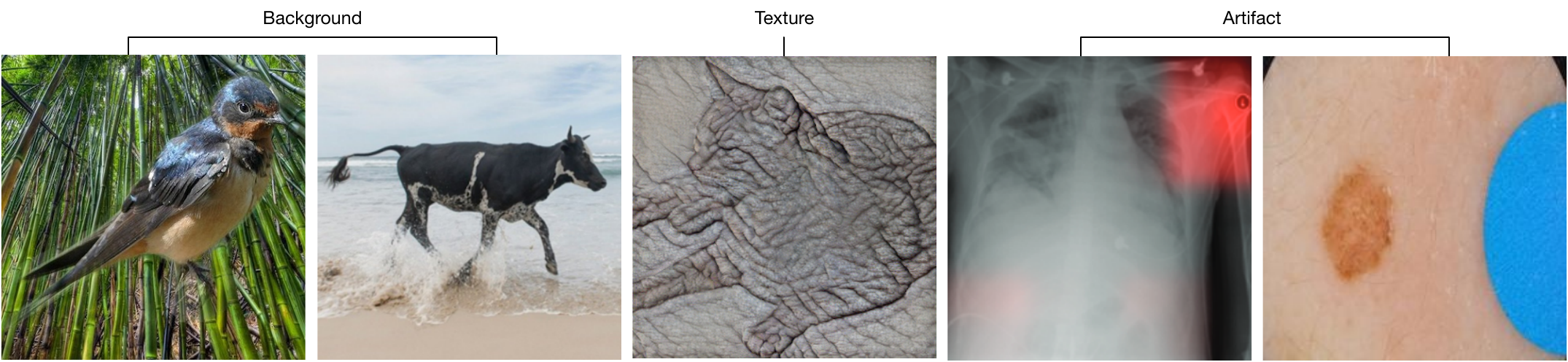}
    \caption{Examples of variety shortcut types including backgrounds, texture and artifacts. The leftmost image is from the \wb dataset, the three center images are from \cite{Geirhos:2020:nature}, and the rightmost image is from the \isic dataset.}
    \label{fig:shortcut-examples}
\end{figure}

Machine learning models in classification tasks tend to learn spurious features that have strong relationships with the target labels are not causal.
Models that rely on spurious correlations for their predictions would, for example, classify a bird as landbird based on the background feature, fail to recognize a cow on the beach, or predict the presence of pneumonia based on background features (see Figure~\ref{fig:shortcut-examples} for some examples). Especially in high-risk domains, such models could have serious consequences: a CNN predicting skin cancer using the presence of a color calibration patch (see Figure~\ref{fig:shortcut-examples}, rightmost image) as a spurious feature fails to detect 68\% of malignant cases when the color patch is absent~\cite{Nauta2021-shortcutlearning}.

During training, models heavily influenced by spurious correlations tend to learn these relationships and memorize samples from so-called \emph{minority} groups where the spurious correlation is not present or which have inverse relationships~\cite{Sagawa:2020b:ICML}. This mechanism helps the models to achieve high average performance during training but leads to poor generalization on the minority group. Therefore, the key objective of methods that mitigate spurious correlations is to improve the performance of minority groups.

Methods such as deep feature re-weighting (DFR)~\cite{Kirichenko:2023:ICLR} and others~\cite{Hakemi:2025,Qiu:2023:ICML} have successfully improved the performance of minority groups without extensive training to minimize group loss, even when models are trained in the presence of spurious correlations. These approaches assume that machine learning models are able to learn sufficient information about all features. Thus, by adjusting only the weights of the classification layer while leaving the learned representation unchanged, the performance of the minority group can be improved. On the other hand, other work~\cite{Le:2023:llr} shows that DFR~\cite{Kirichenko:2023:ICLR} works similarly to a last-layer pruning method, removing a large fraction of neurons that encode spurious features. However, even after re-weighting the classifier weights through DFR, the model still retains spurious information. The analysis in~\cite{Le:2023:llr} provides some initial evidence that spurious features are not completely disentangled in the last layer, but a systematic analysis of how and where spurious features are encoded within models is still lacking.

In this work, we complete the investigation of the learning of spurious features in vision models by analyzing the phenomenon in both CNNs and ViTs models. We re-confirm the influence of spurious correlations and imbalanced data distribution. Further, we investigate the influence of spurious features within the networks, starting from the learned representation space and going deeper into the neurons and components of the models.  We show the limitations of the underlying assumptions of existing spurious mitigation work and explain why they work and how they might fail.
Specifically, our contributions are:
\begin{enumerate}
    \item We show that both ViTs and CNNs learn spurious features and that this behavior can be explained by the representations in latent space (Section~\ref{sec:sc-learning}).
    \item We introduce the neuron spurious score (\sscore), an XAI-based metric to measure a neuron's reliance on spurious features (Section~\ref{sec:setup}).
    \item We show that the level of spurious feature disentanglement in neurons within the latent space differs between CNNs and ViTs. In CNNs, some neurons exclusively encode spurious features, while others encode both spurious and core features (Section~\ref{sec:cnn}). In contrast, in ViTs it is more difficult to find a clear set of neurons only encoding spurious features (Section~\ref{sec:vit}).
    \item We show that spurious features are encoded in different parts of the network, and that these neurons are not only located in the last layer but also in components of the neural network comprising multiple layers (Section~\ref{sec:cnn} for CNNs and Section~\ref{sec:vit} for ViTs).
\end{enumerate}

Our results provide evidence that unlearning spurious correlations is a complex task. While identifying and pruning spurious encoding components can be effective, it may not be sufficient due to the entangled nature of learned representations. Furthermore, pruning methods need to account for architecture-specific aspects, as spurious features are encoded differently in CNNs and ViTs.

The structure of this paper is as follows. First, we introduce the notion of spurious features (Section~\ref{sec:background}). We then describe the general setup of our experiments to analyze spurious correlations and introduce the \sscore as an XAI-based inspection criterion (Section~\ref{sec:setup}).
Section~\ref{sec:sc-learning} empirically shows that CNNs and ViTs are prone to learning spurious features and provides evidence that this is due to data manifold in representation space. Section~\ref{sec:cnn} and Section~\ref{sec:vit} extend the analysis of representation space to all layers of neural networks using inspection techniques specific to CNNs and ViT, respectively. We discuss the main results in Section~\ref{sec:discussion}, review related work in Section~\ref{sec:related-work} and conclude in Section~\ref{sec:conclusion}.

%% file: 02_background.tex
\section{Background} \label{sec:background}

Spurious features $\mathcal{S}$ refer to statistically informative features that do not have a causal relationship with the target labels $\mathcal{Y}$~\cite{Sagawa:2020a:ICLR,Geirhos:2020:nature}. Models that learn spurious features often achieve impressive accuracy on the training dataset by exploiting spurious correlations present in the training data. In a dataset containing target labels $\mathcal{Y}$ and spurious features $\mathcal{S}$, we partition the data into groups based on the combination of labels and spurious features, denoted as $\mathcal{G} = \mathcal{Y} \times \mathcal{S}$. In general, within $\mathcal{G}$ of the training set, there exists at least one group that significantly has smaller size than others and does not contain the corresponding spurious features to the label, referred to as the \emph{minority group}. Models that learn spurious correlations usually fail to predict this particular group during test time.
For example, in the task of classifying bird types, where bird types are highly correlated with the background scene in the images, models tend to learn easier background features rather than bird characteristics (cf. Figure~\ref{fig:shortcut-examples}, leftmost image). A model that predicts based on background features performs well on the training (and i.i.d. test) data since most labels align with the background, and the model only needs to ignore or memorize a few remaining samples from \emph{minority group}~\cite{Sagawa:2020b:ICML}. However, such models fail to generalize to birds on other backgrounds during test time.

Some types of spurious features can be easy to detect for humans, such as background and color~\cite{Sagawa:2020a:ICLR}, or artifacts in domain-specific settings~\cite{Nauta2021-shortcutlearning}. However, some spurious features, such as texture or frequency patterns, may be imperceptible to the human eye~\cite{Geirhos:2018:ICLR,Lin:2023:neurips}. 
Figure~\ref{fig:shortcut-examples} shows examples of shortcuts\footnote{Following related work, we use the terms `spurious features' and `shortcuts' interchangably }.

%% file: 03_experimental-setup.tex
\section{Experimental Setup}
\label{sec:setup}

In this study, we analyze the robustness of various models to spurious correlations. 
In this section, we provide details on the datasets, the vision models, and the evaluation metrics.

\input{Tables/03_fig-dataset}

\subsection{Datasets} 
We consider two datasets: \wb ~\cite{Sagawa:2020a:ICLR} and \isic ~\cite{isic2019}. Examples and data distribution for both datasets are presented in Table~\ref{tab:dataset}.

\paragraph{\wb} is a benchmark dataset for studying spurious correlations in learning. The task is to classify birds as either \emph{water birds} or \emph{land birds}. It is an artificially constructed dataset where bird images from the \textsc{CUB} dataset~\cite{Wah:2011:cub-dataset} are placed onto backgrounds from the \textsc{Places-365} dataset~\cite{Zhou:2018:place-dataset}. To introduce spurious correlations, in the original training set, 95\% of water birds are placed on water backgrounds, and 95\% of land birds are placed on land backgrounds. However, in the test set, this ratio is balanced to evaluate model generalization. The four groups of the dataset are denoted as $\mathcal{G}_0$, $\mathcal{G}_1$, $\mathcal{G}_2$ and $\mathcal{G}_3$ with two minority groups $\mathcal{G}_1$ and $\mathcal{G}_2$ as in Table~\ref{tab:dataset}.

\paragraph{\isic} is a real-world data set for skin cancer detection. The data are obtained from the official website~\footnote{\url{https://www.isic-archive.com}} and labeled as either \emph{benign} or \emph{malignant}. Prior studies~\cite{Rieger:2020:ICML} suggest that the dataset may contain several spurious correlations, such as color patches, rulers or surgical marks, black borders, etc. In this work, we focus on the color patch feature, which appears exclusively in the benign class in nearly 50\% of cases. For evaluation, we construct an artificial test set where color patches are inserted to balance the spurious correlation. The four groups of the dataset are denoted as BwoP (benign without patch), BwP (benign with patch), MwoP (malignant without patch) and MwP (malignant with patch), with MwP representing the minority group. 

\subsection{Models and Hyperparameters} 
We evaluate both convolutional neural networks (CNN) and Vision Transformer (ViT) models.  As representatives for CNNs, we use differently sized Resnet models, namely ResNet-18, ResNet-50~\cite{He:2016:resnet}, and ResNeXt~\cite{Xie:2017:CVPR}. We evaluate two different  ViT, namely ViT-B/16~\cite{Dosovitskiy:ICLR:2021}, and DeiT~\cite{Touvron:2021:ICML}. 
All models use pre-trained weights from ImageNet-1K~\cite{imagenet:2015}.
We do not adjust the hyper-parameters to optimize the worst-group accuracy. We follow the finetuning method of previous work~\cite{Kirichenko:2023:ICLR,Sagawa:2020a:ICLR} and use the following hyperparameter settings: each model is trained for 100 epochs with learning rate 0.001, weight decay $10^{-4}$ and SGD optimizer~\cite{Robbins:1951:stochastic}. We adapt the batch size to the image input size to accommodate memory size and use 32 for \wb and 64 for \isic.

\subsection{Evaluation Metrics}
In our experiments, we use the standard metrics for evaluating reliance on spurious correlation: average accuracy and worst-group accuracy. To quantify the reliance of single neurons on a spurious input feature, we introduce the \sscore.  

\paragraph{Worst-group and Average Accuracy.}
In general, a model's robustness is measured by average accuracy — the proportion of correct classifications out of all predictions.
Models that exploit spurious correlations often achieve high average accuracy (AVG) but perform significantly worse for a particular group within $\mathcal{G}$. The accuracy of this group is called worst group accuracy (WGA). Additionally, we denote the difference between AVG and WGA as GAP. A smaller GAP indicates greater robustness to spurious correlations.
\input{Tables/03_fig-segmented-mask}

\paragraph{Neuron Spurious Score.} In Section~\ref{sec:cnn} and Section~\ref{sec:vit}, we investigate whether there exist neurons in the penultimate layer that purely encode the spurious features. To measure the extent to which a neuron focuses on the spurious region in the input, we introduce the \sscore. We consider a model $f$ constructed by a feature extractor $f_{\text{enc}}: \mathcal{X} \rightarrow \mathbb{R}^d$ and a linear classification layer $h: \mathbb{R}^d \rightarrow \mathcal{Y}$. Given the input $x_j$, its corresponding penultimate representation $f_{\text{enc}}(x_j)=z_j \in \mathbb{R}^d$ and the spurious segmentation $m_j$ given by a binary matrix (e.g., patch segmentation in ISIC), we use GradCAM~\cite{gradcam:2020:ComVis} to compute the heatmap attribution over input, denoted as $a_j^i$, from neuron $i$ of $z_j$. 
To emphasize the highly focused region identified by neuron $i$, we set a threshold $\alpha$ and binarize the heatmap $a_j^i$ into $b_j^i$. The neuron spurious score, \sscore, measures the proportion of the neuron’s focusing region that corresponds to the region of the spurious feature. The \sscore of neuron $i$ is calculated by averaging over $N$ samples
$$s^i = \frac{1}{N} \sum_{j=1}^N\frac{\sum b_j^i \odot m_j}{\sum b_j^i}$$
The neuron \sscore $s^i$ ranges between $[ 0,1 ]$, where $s^i = 0$ indicates that either the neuron does not activate any input region or none of the focus regions overlap with the spurious segmentation.
We consider a neuron with a high \sscore a spurious feature-encoding neuron in the representation space. Examples of the input image, the segmented mask, the heatmap of a neuron and the  \sscore attribution of a single neuron are shown in Figure~\ref{fig:mask}.

%% file: Tables/03_fig-dataset.tex
\begin{table}[t]
    \centering
    
    \caption{Overview of the \wb and \isic datasets. Percentages in \textbf{\#~Train} presents the proportion of that group within the class $\mathcal{Y}$.} \label{tab:dataset}
    \renewcommand{\arraystretch}{1.5}
    \setlength{\tabcolsep}{3pt}
    \newcolumntype{C}[1]{>{\centering\arraybackslash}p{#1}}
    \begin{tabular}{c c c c c}
        \toprule
        \textbf{\wb} & $\mathcal{G}_0$ & $\mathcal{G}_1$ & $\mathcal{G}_2$ & $\mathcal{G}_3$ \\
        \midrule
        
        & \includegraphics[width=2cm]{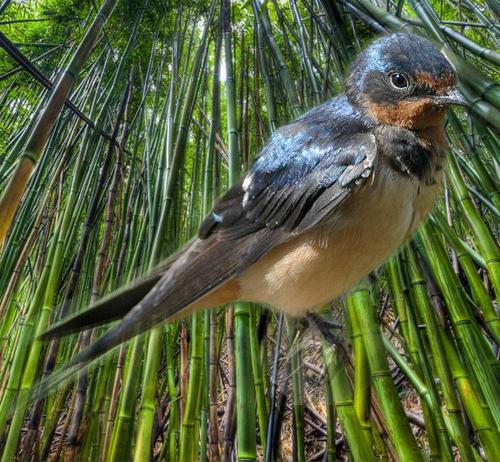}  
        & \includegraphics[width=2cm]{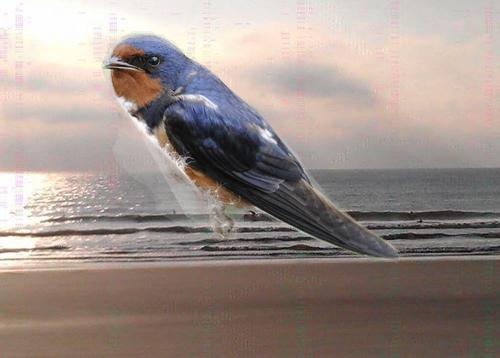}  
        & \includegraphics[width=2cm]{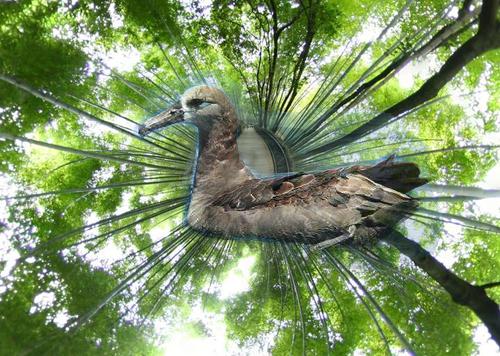}  
        & \includegraphics[width=2cm]{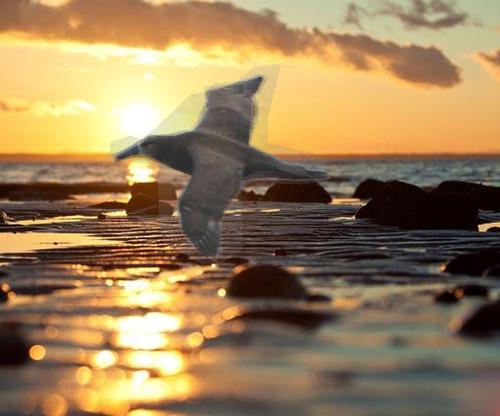}  \\
        $\mathcal{Y}$ & landbird & landbird & waterbird & waterbird \\
        $\mathcal{S}$ & land & water & land & water \\
        \midrule
        \textbf{\# Train} & 3,518 (95\%) & 185 (5\%) & 55 (5\%) & 1,037 (95\%) \\
        \textbf{\# Test} & 2,255 & 2,255 & 642 & 642 \\
    \end{tabular}
    
    
    \begin{tabular}{c c c c c}
        \toprule
        \isic & \maroon{BwoP} & \teal{BwP} & MwoP & MwP \\
        \midrule
        & \includegraphics[width=2cm]{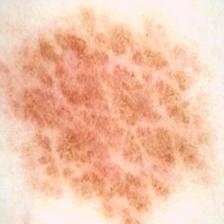}  
        & \includegraphics[width=2cm]{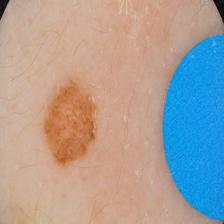}  
        & \includegraphics[width=2cm]{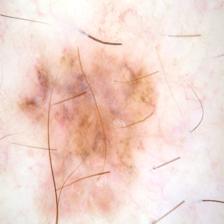}  
        & \includegraphics[width=2cm]{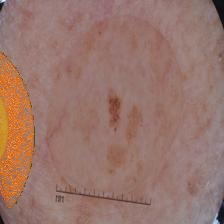}  \\
        $\mathcal{Y}$ & benign & benign & malignant  & malignant \\
        $\mathcal{S}$ & no patch & patch & no patch & \emph{inserted} patch \\
        \midrule
        \textbf{\# Train} & 6,314 (53\%) & 5,526 (47\%) & 1,571 (100\%) & 0 (0\%) \\
        \textbf{\# Test} & 3,158 & 2,763 & 821 & 821 \\
        \bottomrule
    \end{tabular}

\end{table}

%% file: Tables/03_fig-segmented-mask.tex
\begin{figure}[t!]
\includegraphics[trim={0 0cm 0 0cm},clip,width=0.4\textwidth]{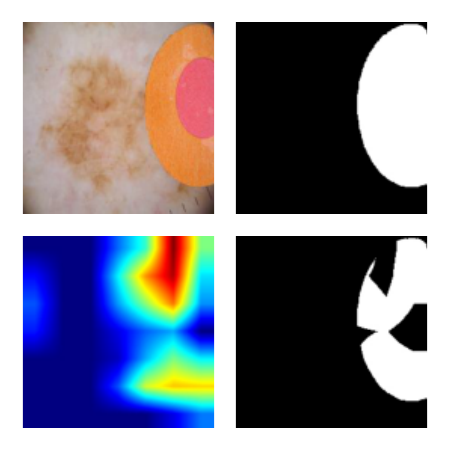}
\hspace{10mm}
\includegraphics[trim={0 0cm 0 0cm},clip,width=0.4\textwidth]{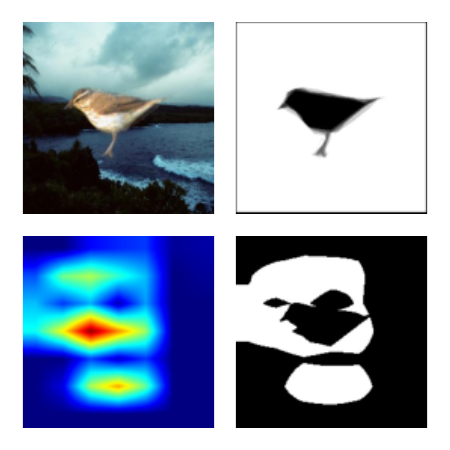}
\centering
\caption{
Visualization of the \sscore derivation by examples on \isic dataset (left) and \wb (right). A binary mask of the spurious feature (color patch or background) is obtained from the annotations in the training data set (top row).  We use GradCam to obtain a feature attribution heatmap from a neuron and  intersect its binarized version with  the segmentation map (bottom row).}
\label{fig:mask}
\end{figure}

%% file: 04_learning.tex
\section{Learning Spurious Features} 
\label{sec:sc-learning}
In this section, we investigate the extent to which convolutional neural networks (CNNs) and vision transformers (ViTs) are susceptible to learning spurious correlations. 
First, we analyze the performance of the models for different groups in the data and with varying ratios of spurious correlations (Section~\ref{ssec:learning:performance}). 
Second, we investigate whether this behavior can be explained by the representations in latent space (Section~\ref{ssec:learning:representation}). 

\subsection{Performance on Groups}
\label{ssec:learning:performance}
We fine-tune pre-trained CNNs and ViTs models on two datasets: \wb and \isic. The effect of spurious features is shown by the difference between the worst-group accuracy and the average accuracy during test time. Table~\ref{tab:modelzoo} shows the testing performances across different models. Overall, all vision models tend to learn spurious correlations as shown by the significant gap (GAP) between the average accuracy (AVG) and the worst-group accuracy (WGA).
\input{Tables/04_tab-WGA-modelzoo}

To analyze the extent to which models react to the presence of spurious correlations, we train a \rn by varying the proportion of minority groups in the \wb dataset during training (cf. Table~\ref{tab:wb-spurious-ratio}). A minority ratio of 50\% means that there is no spurious correlation in this modified dataset, and a ratio of 0\% means that there is no minority sample in the training set. 
In the test set, we take the same number of samples from each group (i.e., a balanced subset). The two groups with an identical class label share the same foreground images, while the background is either water or land. Table~\ref{tab:wb-spurious-ratio} shows that the performance of the minority group gets worse as its proportion in the training data decreases (cf. $\mathcal{G}_1$ and $\mathcal{G}_2$), while the majority group performance remains nearly unchanged and consistently high — above 90\% and even higher than when training without spurious correlations. 
\input{Tables/04_tab-wb-spurious-ratio}
However, even in the worst-case scenario (0\% minority groups in the training set), the models can still correctly predict out-of-distribution (O.O.D.) samples in the test set, though with less than 50\% accuracy. This shows that models not only learn spurious features but also capture core features.

\paragraph{Takeaways.} Vision models are susceptible to spurious correlations but still retain generalization ability.

\subsection{Analyzing Latent Space} 
\label{ssec:learning:representation}
\input{04-fig-representations}

We analyze the feature representations from the penultimate layer of the training data set of both CNN (\rn) and ViT (\vit). To visualize all data points, we show a t-SNE~\cite{Van:2008:tSNE} projection of the output from the feature extractor $f_{\text{enc}}$ of each model. In the ViT visualizations (Figure~\ref{fig:representation}, right column), both, \isic and \wb show a clear trend that samples with the same spurious features across classes blend in the representation manifold (BwoP and MwoP in \isic; waterbird on water and landbird on water in \wb). In the \rn representation, we observe that even the boundaries between classes are separated in both cases, within each class, the data clusters according to spurious features. Note that in all cases, the models achieve close to 99\% training accuracy. We hypothesize that depending on the model architecture, the spurious features are learned differently, however, in all cases, those features are well recognized and have a high impact on the classification result.
\paragraph{Takeaways.} In both \rn and ViT-B/16, we observe an identical phenomenon that samples with the same spurious features tend to lie closer to each other in the representation space, even with high training accuracy.

%% file: Tables/04_tab-WGA-modelzoo.tex
\begin{table}[t!bh]
    \centering
    \caption{Average (AVG) and worst-group accuracy (WGA) across 5 different runs on the test set for both CNN and ViT models trained on \wb and \isic. Showing mean and standard deviation.}
    \label{tab:modelzoo}
    \setlength{\tabcolsep}{6pt}
     \begin{tabular}{lcccccc}
    \toprule
         & \multicolumn{3}{c}{\wb}  &  \multicolumn{3}{c}{\isic} \\
         & AVG & WGA$\uparrow$ & GAP$\downarrow$ & AVG & WGA$\uparrow$ & GAP$\downarrow$\\
         \midrule
       \rn  & 0.83 {\tiny $\pm$ 0.01} & 0.46 {\tiny $\pm$ 0.02} & 0.37 & 0.84 {\tiny $\pm$ 0.01} & 0.37 {\tiny $\pm$ 0.01} & 0.47 \\
       ResNet50  & 0.88 {\tiny $\pm$ 0.00} & 0.63 {\tiny $\pm$ 0.02} & 0.25 & 0.83 {\tiny $\pm$ 0.01} & 0.22 {\tiny $\pm$ 0.01} & 0.61 \\
       ResNeXt  & 0.89 {\tiny $\pm$ 0.00} & 0.70 {\tiny $\pm$ 0.02} & 0.19 & 0.86 {\tiny $\pm$ 0.01} & 0.42 {\tiny $\pm$ 0.01} & 0.44 \\
       ViT-B/16  & 0.87 {\tiny $\pm$ 0.01} & 0.65 {\tiny $\pm$ 0.01} & 0.22 & 0.83 {\tiny $\pm$ 0.01} & 0.16 {\tiny $\pm$ 0.01} & 0.67 \\
       DEiT  & 0.88 {\tiny $\pm$ 0.00} & 0.66 {\tiny $\pm$ 0.01} & 0.22 & 0.83 {\tiny $\pm$ 0.01} & 0.12 {\tiny $\pm$ 0.01} & 0.71 \\
       \bottomrule
    \end{tabular}
\end{table}

%% file: Tables/04_tab-wb-spurious-ratio.tex
\begin{table}[tbh]
    \centering
    \caption{Influence of the correlation ratio of the spurious feature.   \rn on \wb. We keep the number of samples in the training set fixed (same as in the original dataset) and vary the percentage of the minority groups $\mathcal{G}_1$ and $\mathcal{G}_2$ within its corresponding class (originally 5\%). In the test set, the two groups in each pair ($\mathcal{G}_0$, $\mathcal{G}_1$) and ($\mathcal{G}_2$, $\mathcal{G}_3$) share the same foreground images (the birds) while adapting to different backgrounds (water or land).}
    \label{tab:wb-spurious-ratio}
    \setlength{\tabcolsep}{12pt}
\begin{tabular}{lccccccc}
    \toprule
        Minority Ratio & $\mathcal{G}_0$ & $\mathcal{G}_1$ & $\mathcal{G}_2$ & $\mathcal{G}_3$ & AVG & GAP$\downarrow$\\
         \midrule
         50\% & 0.98 & 0.98 & 0.84 & 0.83 & 0.91 & 0.08 \\
         25\% & 0.99 & 0.95 & 0.74 & 0.90 & 0.94 & 0.20 \\
         5\% & 0.99 & 0.78 & 0.46 & 0.91 & 0.84 & 0.38 \\
         0\% & 0.99 & 0.34 & 0.22 & 0.94 & 0.65 & 0.43 \\
       \bottomrule
    \end{tabular}
\end{table}

%% file: 04-fig-representations.tex
\begin{figure}[htb]
    \centering
    \begin{subfigure}{0.49\textwidth}
        \includegraphics[width=\textwidth, trim=1.5cm 1.5cm 0.5 1.5cm, clip]{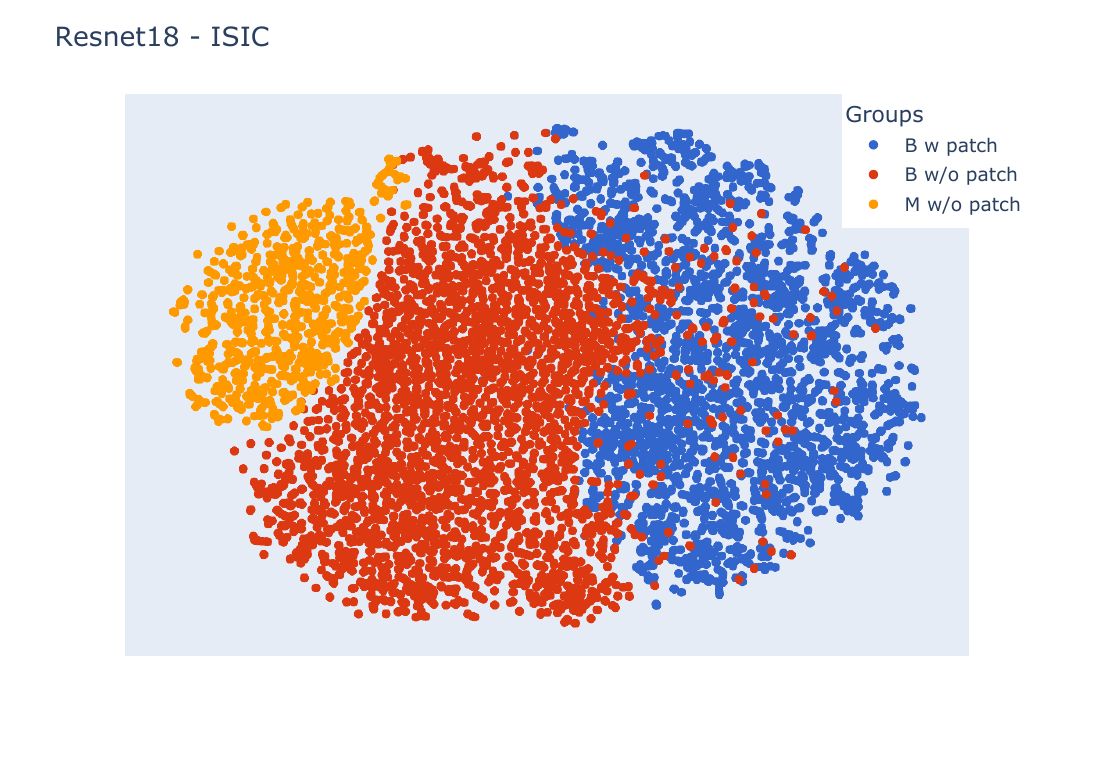}
        \caption{ResNet18 - ISIC}
    \end{subfigure}
    \hfill
    \begin{subfigure}{0.49\textwidth}
        \includegraphics[width=\textwidth, trim=1.5cm 1.5cm 0.5 1.5cm, clip]{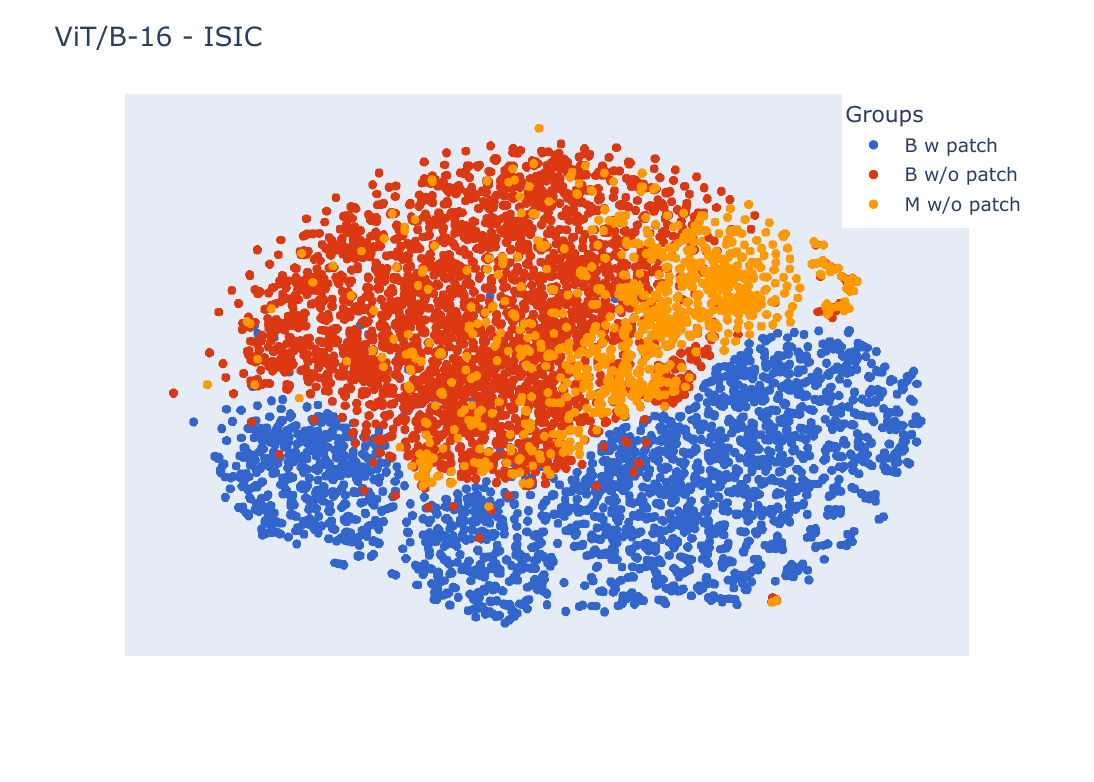}
        \caption{ViT/B-16 - ISIC}
    \end{subfigure}\\
        \begin{subfigure}{0.49\textwidth}
        \includegraphics[width=\textwidth, trim=1.5cm 1.5cm 0.5 1.5cm, clip]{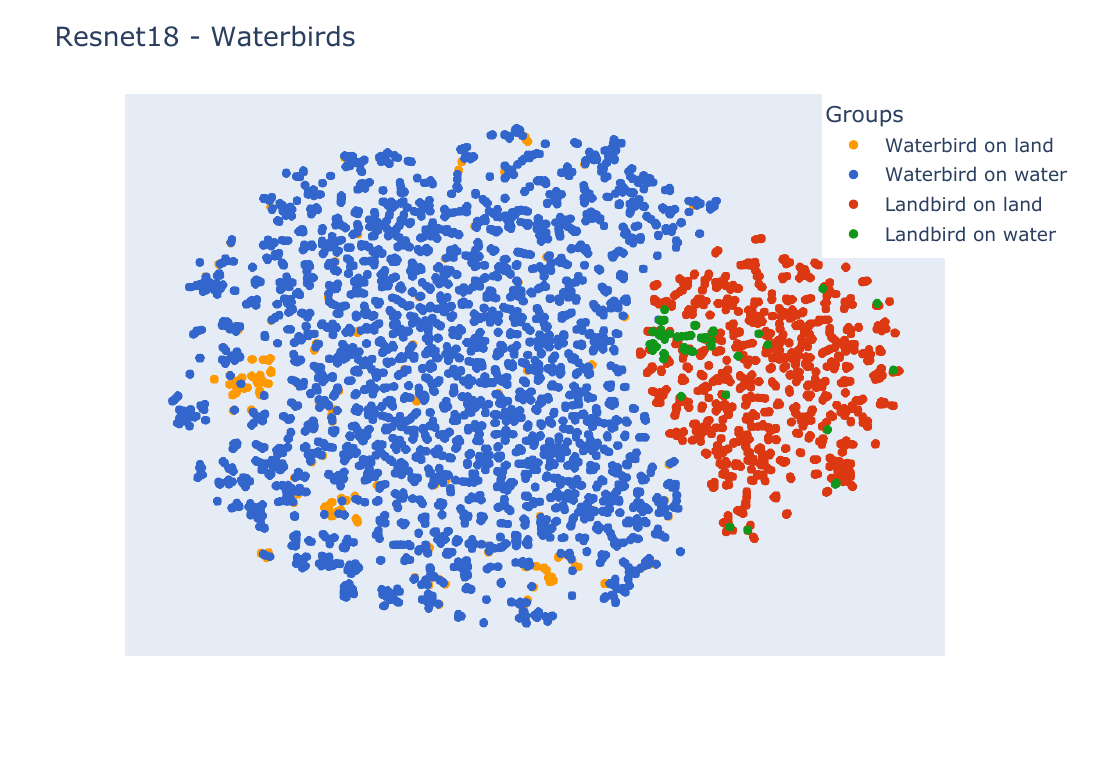}
        \caption{ResNet18 - \wb}
    \end{subfigure}
    \hfill
    \begin{subfigure}{0.49\textwidth}
        \includegraphics[width=\textwidth, trim=1.5cm 1.5cm 0.5 1.5cm, clip]{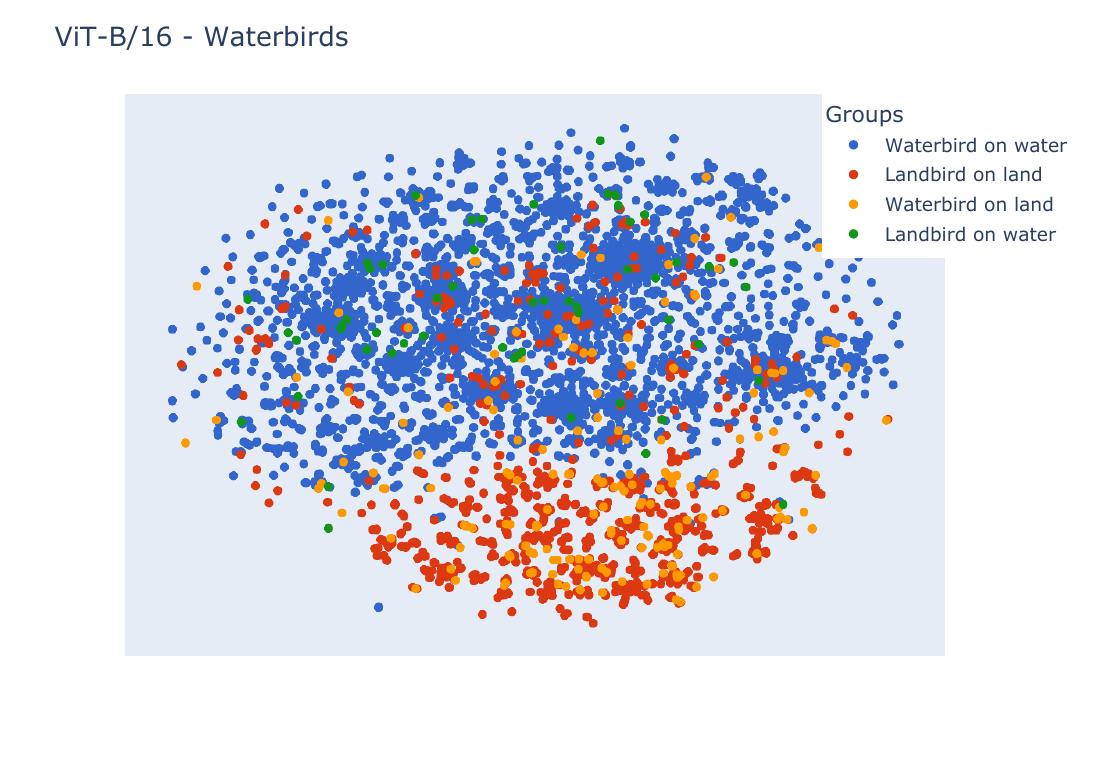}
        \caption{ViT/B-16 - \wb}
    \end{subfigure}
\caption{Visualization of representation of last layers. Clusters are mainly defined by the spurious attribute (patch for ISIC, land/water for birds) and not by the classes (malignant/benign for ISIC and landbird/waterbird for birds).}
\label{fig:representation}
\end{figure}

%% file: 05-1_cnn-disentanglement.tex
\section{Encoding of Spurious Features in CNNs}
\label{sec:cnn}

In Section~\ref{sec:sc-learning} we showed that the latent representations in the last layer of ViT and CNNs are governed by spurious features, i.e., clusters in latent space are defined more by the spurious features than by the target labels. In this section, focusing on CNNs, we analyze the extent to which this behavior can be attributed to individual neurons in different neural network layers. We begin by examining the disentanglement and entanglement of neurons in the penultimate layer (Section~\ref{ssec:cnn:disentanglement} and Section~\ref{ssec:cnn:entanglement}).
For CNN-based models, we we investigate whether models learn disentangled information for different data groups (Section~\ref{ssec:cnn:modular}) using techniques for subnetwork extraction (so called network modulars)~\cite{csordas:2020:ICLR}.

\subsection{Neuron Disentanglement} 
\label{ssec:cnn:disentanglement}
\input{Tables/05-1_fig-sscore-heatmap}
We analyze single neurons in the last layer to test whether any of them are highly related to the spurious region. In each dataset, we compute the neuron spurious score \sscore (cf. Section~\ref{sec:setup}) over 50 random training samples.

On \isic, the \sscore of neurons ranges from 0.0 to 0.8. Visualization of three neurons from different score ranges is shown in Figure~\ref{fig:sscore-heatmap}. We determine the \sscore ranges based on the proportion of the heatmap that overlaps with the mask segmentation of spurious features. A neuron receives a low \sscore if on average less than 20\% of the heatmap overlaps with the spurious mask (\sscore$< 0.2$), and a high \sscore if more than 70\% of the heatmap focuses on the spurious region (\sscore$>0.7$). Otherwise, the neuron receives a mid-range \sscore.  For neurons with high or low \sscore, the main focus region (more red areas) consistently highlights either the patch (spurious feature) or the lesion (core feature). Meanwhile, the neurons in mid-range \sscore shift their focus between spurious and core features depending on the sample.

\begin{table}[tb]
    \centering
    \caption{Group accuracy of ISIC after pruning some sets of neurons in the last layer based on the neuron spurious score (\sscore).}
    \label{tab:prune-last-layer}
    \begin{tabular}{p{0.3\textwidth}p{0.1\textwidth}p{0.1\textwidth}p{0.1\textwidth}p{0.1\textwidth}p{0.08\textwidth}}
    \toprule
         & BwoP & BwP & MwoP & MwP & AVG \\
         \midrule
       \rn  & 0.92 & 0.99 & 0.57 & 0.37 & 0.84\\
       Pruning (\sscore > 0.7)  & 0.90 & 1.00 & 0.59 & 0.40 & 0.85\\
       \midrule
       \multicolumn{6}{c}{Fine-tune last-layer weights with group balanced set} \\
       \rn  & 0.81 & 0.99 & 0.75 & 0.51 & 0.84\\
       Pruning (\sscore > 0.7)  & 0.81 & 0.99 & 0.77 & 0.56 & 0.85\\
       \bottomrule
    \end{tabular}
\end{table}

\paragraph{Influence of Spurious-encoding Neurons.} By setting the weights connecting spurious-encoding neurons to the classes to zero, we observe a slight improvement in the WGA (Table~\ref{tab:prune-last-layer}, rows 1–2).\footnote{Here, we set the weights to zero without any retraining} To reduce the influence of a highly imbalanced data distribution on the classifier, we additionally fine-tune the linear classifier using a group-balanced dataset in two cases: baseline \rn and \rn with spurious-encoding neurons deactivated (Table~\ref{tab:prune-last-layer}, rows 3-4). Fine-tuning after pruning some spurious-encoding neurons results in a slightly higher improvement in WGA compared to fine-tuning the baseline, indicating that the pruned representation better captures invariant features. 
From these experiments, we conclude that even though we can find some critical neurons (using labels and annotations of the spurious feature), we do not know how neurons interact with each other and, therefore, can not find a complete set of neurons that encode spurious features.

\paragraph{Takeaways.} Our analyses in this section show that spurious features are represented and to some extend disentangled in the representation space and simply turning off those neurons improves robustness to spurious correlations.

%% file: Tables/05-1_fig-sscore-heatmap.tex
\begin{figure}[ht]
\centering

\includegraphics[width=\textwidth]{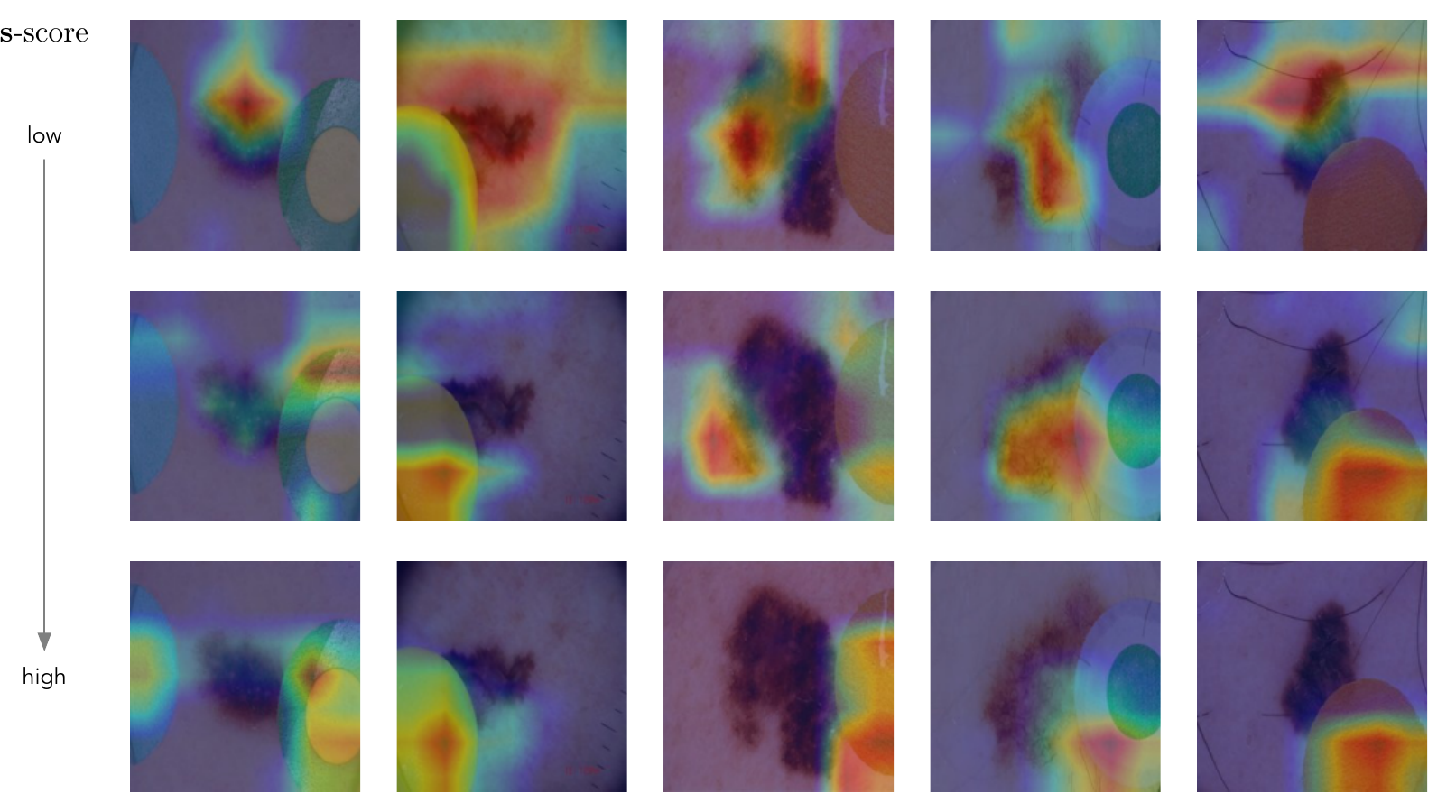}
\caption{Neuron heatmaps overlaid with original images, illustrating the activation of three different neurons (rows) across five sample images (columns). }
\label{fig:sscore-heatmap}
\end{figure}


%% file: 05-2_cnn-entanglement.tex
\subsection{Neuron Entanglement} \label{ssec:cnn:entanglement}
Also in the direction of not retraining feature extractor $f_{\text{enc}}$, deep feature re-weighting (DFR) method~\cite{Kirichenko:2023:ICLR} and the subsequent analysis~\cite{Le:2023:llr} suggest that using a group-balanced dataset to select essential neurons in the representation space and disabling all other neurons might make models more robust to spurious correlations.
\paragraph{DFR Method.} DFR keeps the learned representation of trained models unchanged, retraining only the classification layer with a group-balanced validation set using logistic regression. The logistic regression hyperparameters are optimized for group performance with another group-balanced dataset. Under the assumption that trained models learn sufficient information despite the existence of spurious correlations, this approach uses a group-balanced set to seek the optimal neuron combination that is not influenced by spurious correlations.
\begin{table}[tb]
    \centering
    \caption{Performance of applying DFR on \rn for two datasets \isic and \wb. Prune$_h$ denotes the pruning ratio of the classification layer weights.}
    \label{tab:dfr-wga}
    \begin{tabular}{p{0.18\textwidth}p{0.15\textwidth}p{0.14\textwidth}p{0.14\textwidth}p{0.14\textwidth}p{0.14\textwidth}}
    \toprule
        & Model & WGA & AVG & Prune$_{h}$ & Avg. $\mathbf{s}$-score\\
         \midrule
         \multirow{2}{*}{ISIC} & Baseline & 0.37 &  0.85 & -  & 0.40\\
         & DFR & 0.71 & 0.79 & 80\% & 0.39  \\
       \midrule
         \multirow{2}{*}{\wb} & Baseline & 0.46 & 0.83 & - &  0.45\\
        & DFR & 0.82 & 0.87 & 52\% & 0.42  \\
        \bottomrule
    \end{tabular}
\end{table}

\begin{figure}[tb]
\includegraphics[trim={0 0cm 0 1cm},clip,width=0.4\textwidth]{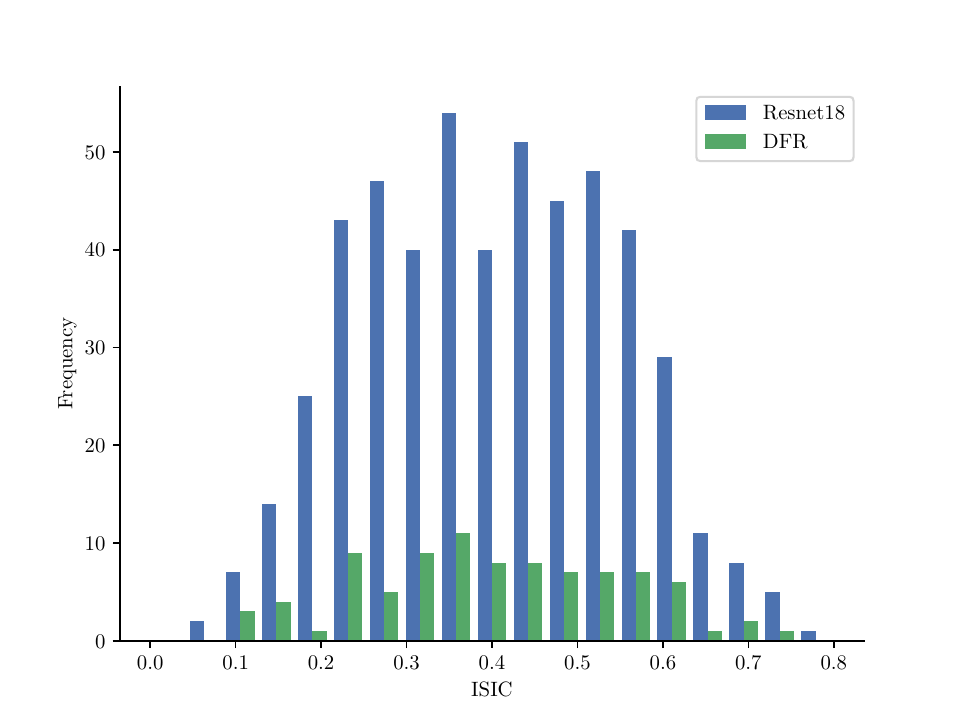}
\includegraphics[trim={0 0cm 0 1cm},clip,width=0.4\textwidth]{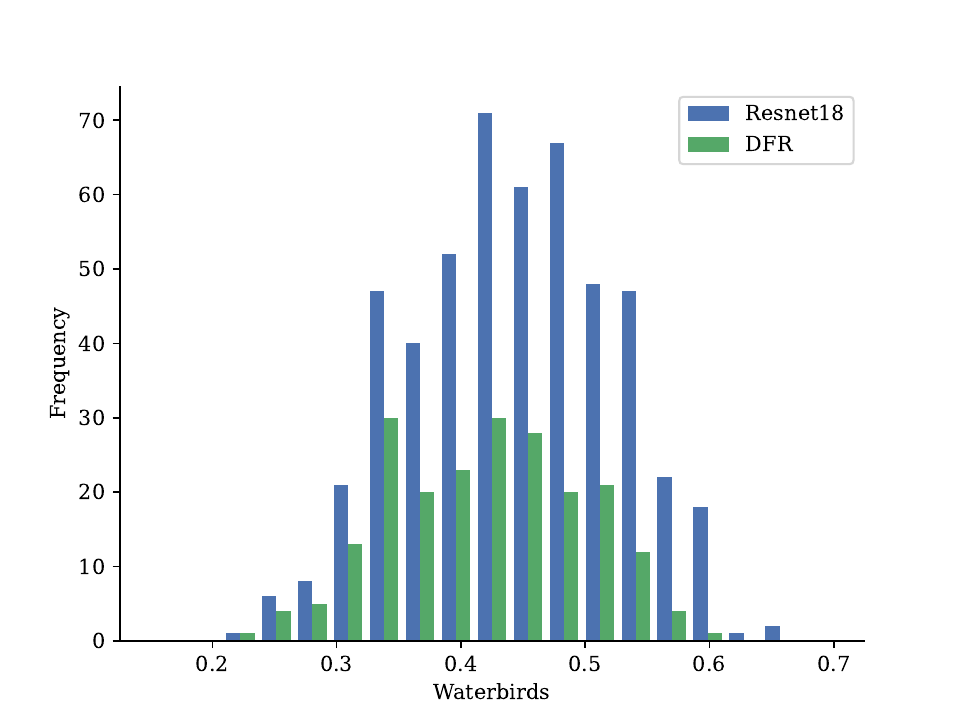}
\centering
\caption{Average \sscore of \rn and DFR trained on \isic (left) and \wb (right).}
\label{fig:s-score}
\end{figure}
Table~\ref{tab:dfr-wga} shows the effectiveness of the DFR method on CNN models trained with \isic.
DFR significantly improves the WGA while zeroing a large number of weights in the classification layer. This means that there is only information encoded in a small number of neurons of the embedding layer that is necessary for classification, and removing them makes models more robust. However, contrary to the naive approach of eliminating neurons that are strongly focused on spurious regions (cf. Table~\ref{tab:prune-last-layer}), our analysis shows that DFR removes a large number of unnecessary neurons while maintaining the same \sscore distribution (cf. Table~\ref{tab:dfr-wga} and Figure~\ref{fig:s-score}). This indicates that DFR does not change the \emph{diversity} of the learned features, whether they are spurious or not. 
With a similar underlying hypothesis, the method of~\cite{Hakemi:2025} also uses both learned core and spurious features in the latent space. This approach succeeds in improving the WGA by searching for a single weight of the classification layer that most activates for the minority group and editing only that weight. 

We hypothesize that instead of genuinely selecting core features encoded in the representation space, the effectiveness of these classifier adaptation methods comes from learning new classifier weights to fit the new non-spurious data (group-balanced). This suggests that interactions between neurons strongly influence classification and that optimizing a subset of neurons and connection weights can improve group-specific performance. However, in alignment with the findings in~\cite{Le:2023:llr}, we conclude that these approaches do not truly eliminate the learned spurious correlations.

\paragraph{Takeaways.} 
Without adapting the learned information, but the interaction between the representation neurons, we can significantly improve the performance of a particular group. However, there is no guarantee that the spurious correlations learned by the models will be completely eliminated.

%% file: 05-3_cnn-modular.tex
\subsection{Disentangled Components}
\label{ssec:cnn:modular}
As we showed in Section~\ref{ssec:cnn:entanglement}, retaining neurons with medium or high \sscore while adjusting their influence on the classifier can reduce the impact of spurious feature learning. Therefore, we hypothesize that there are additional conditions or signals earlier in the network that allow the model to use these neurons more effectively in certain cases. In the following, we apply pruning and subnetwork learning to analyze their effect on spurious correlations.

\paragraph{Task-oriented Pruning.}
We investigate whether pruning neurons in deeper network layers can help reduce reliance on spurious features. PruSC~\cite{Le:2024:spuriousity} and DCWP~\cite{Park:2023:CVPR} are two pruning methods designed to mitigate spurious correlations. Notably, both methods prune neurons based on frozen trained weights by learning a mask on the weights, i.e., they refine the learned features instead of re-learning from scratch.
Applying PruSC to \rn with the \isic dataset results in pruning 65\% of the neurons in the last layer and 48\% of the neurons across the entire model. 
Comparing the average \sscore before and after pruning (cf. Figure~\ref{fig:prusc}), we observe that pruning effectively reduces the connections to spurious features. While improving performance for the worst-case group, PruSC yields a significantly lower average \sscore, indicating a shift toward using less spurious features.

\input{Tables/05-2_fig-sscore-prusc}

\paragraph{Group-specific Components Learning.} 
We further investigate how the relation between core and spurious features is encoded and whether those spurious features are learned and disentangled within the network. The ISIC dataset with training set $\mathcal{D}$ contains three groups: \maroon{BwoP}, \teal{BwP}, and MwoP. We train model $f$ fully on $\mathcal{D}$.

\begin{figure}[tb]
\centering
\includegraphics[width=0.7\textwidth]{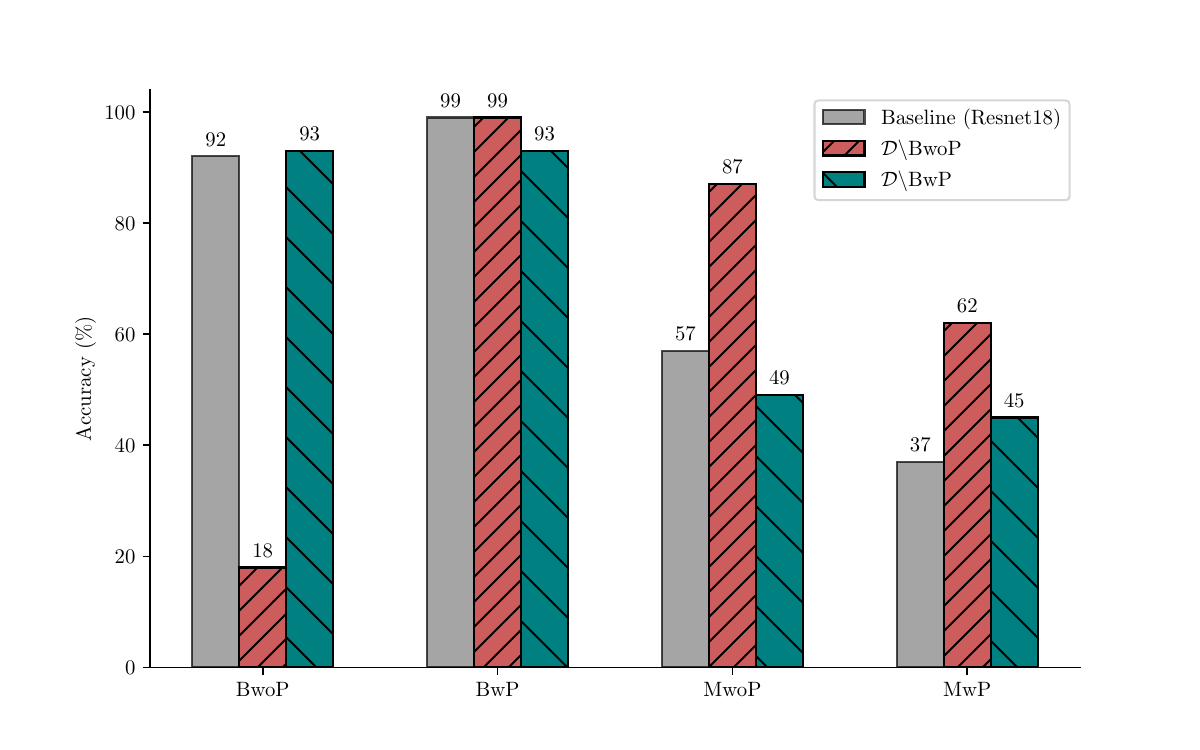}
\caption{Group accuracy of a subnetwork trained with \teal{$\mathcal{D} \setminus \{ \text{BwP} \}$} and \maroon{$\mathcal{D} \setminus \{ \text{BwoP} \}$}.}
\label{fig:modular-isic}
\end{figure}

Applying a similar technique as~\cite{csordas:2020:ICLR}, we freeze all the weights of $f$, and train a binary mask on each weight of $f$ with a subpopulation of the training data. To avoid trivial results in a binary classification task, we can not remove an entire class as in the original paper. However, we hypothesize to obtain a subnetwork that removes all the relevant components that are purely responsible for a specific group. We conduct the study on two sub-dataset: \teal{$\mathcal{D} \setminus \{ \text{BwP} \}$} and \maroon{$\mathcal{D} \setminus \{ \text{BwoP} \}$} by removing the group data \teal{BwP} and \maroon{BwoP} from the training respectively. After pruning the model to 80\% of the total number of weights~\footnote{We ensure that no layer is entirely pruned.}, we evaluate the resulting model with the official test set. The results are shown in Figure~\ref{fig:modular-isic}.

\paragraph{Case 1: Removing \maroon{purely benign} cases.} By removing the entire group \maroon{BwoP}, the accuracy of this group drops significantly from 92\% to 18\%, while the performances of other groups are unchanged or increased. It proves that the group \maroon{BwoP} or the features belonging to the benign class are encoded and disentangled within the network, forming a \maroon{benign-encoding} component. Thus, deleting this component can lead to a significant drop in the performance of a particular group.

\paragraph{Case 2: Removing \teal{patches-containing} cases.} By removing the group \teal{BwP}, we systematically test whether we can eliminate all the \teal{patch-encoding} components within the trained network. The resulting performance after pruning is nearly identical between the two groups within a class (both groups of the benign class obtain 99\% accuracy, and both groups of the malignant class obtain approximately 49\% accuracy). This suggests that the removed connections are indeed responsible for encoding the existence of feature patches. However, we observe that when these connections encoding patches are removed, the performance of group MwoP drops. This means that the removed \teal{patch-encoding} component not only contains information about the patches feature but also important information for predicting the malignant class.

Notably, the group \teal{BwP} can be predicted by using either the \teal{patch-encoding} or the \maroon{benign-encoding} component, and therefore, the group accuracy remains high in both cases.

\paragraph{Takeaways.} We conclude that i) task-oriented pruning is a promising approach for mitigating spurious correlations by turning off connections that contribute to the spurious features, and ii) pruning methods are most effective when models learn features in disentangled subnetworks. 

%% file: Tables/05-2_fig-sscore-prusc.tex
\begin{figure}[htbp]
    \begin{minipage}[b]{0.45\textwidth}
        \centering
        \begin{tabular}{l c c}
            \toprule
            & \rn & PruSC  \\
            \midrule
            WGA  & 0.37 & 0.73 \\
            Avg. \sscore  & 0.40 & 0.18 \\
            Prune$_{h}$ & - & 65\% \\
            Prune$_{f}$ & - & 48\% \\
            \bottomrule
        \end{tabular}
        \vspace{0.8cm}
    \end{minipage}
    \begin{minipage}[b]{0.45\textwidth}
        \centering
        \includegraphics[width=\textwidth]{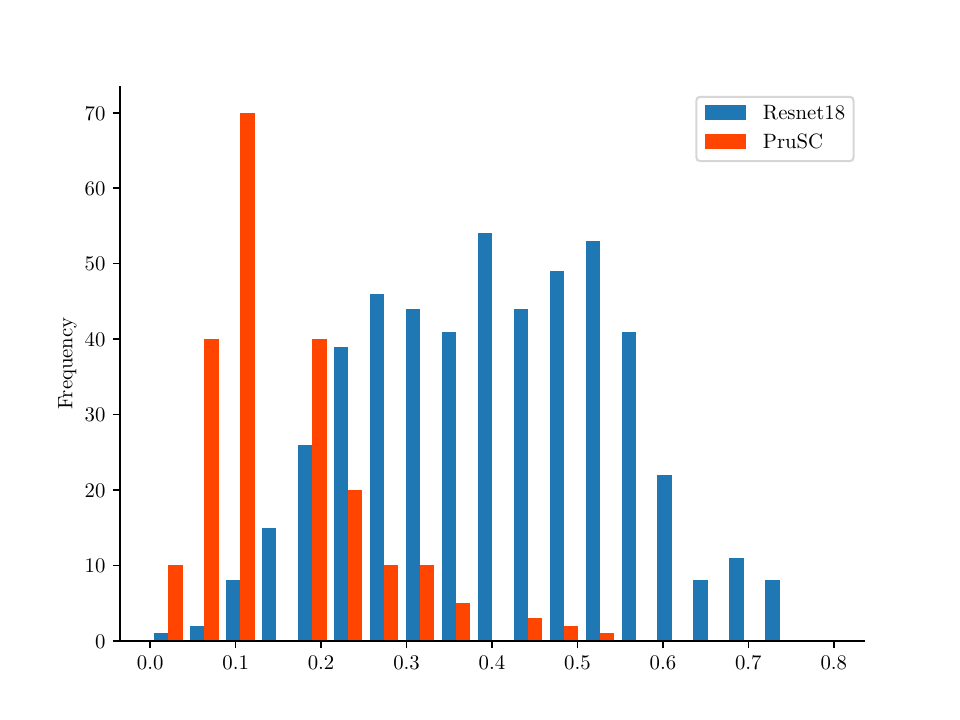}
    \end{minipage}
\caption{\sscore distribution. Prune$_{h}$ denotes the pruning ratio of the classification layer weights, Prune$_{f}$ denotes the pruning ratio of the entire model.}
\label{fig:prusc}
\end{figure}

%% file: 06-1_vit-lastlayer.tex
\section{Encoding of Spurious Features in ViTs} \label{sec:vit}
In this section, we focus on analyzing the entanglement learning of spurious features in ViT models. Following a similar approach to CNNs, we begin by examining neurons in the penultimate layer (Section~\ref{ssec:vit:last-layer}). Next, due to the unique multi-heads self-attention learning mechanism in ViTs, we focus on analyzing spurious features learned in attention heads (Section~\ref{ssec:vit:attention-head}).

\subsection{Last-layer Representation in ViTs} \label{ssec:vit:last-layer}
\begin{figure}[tbh]
\centering
\includegraphics[trim={0 0cm 0 0cm},clip,width=0.4\textwidth]{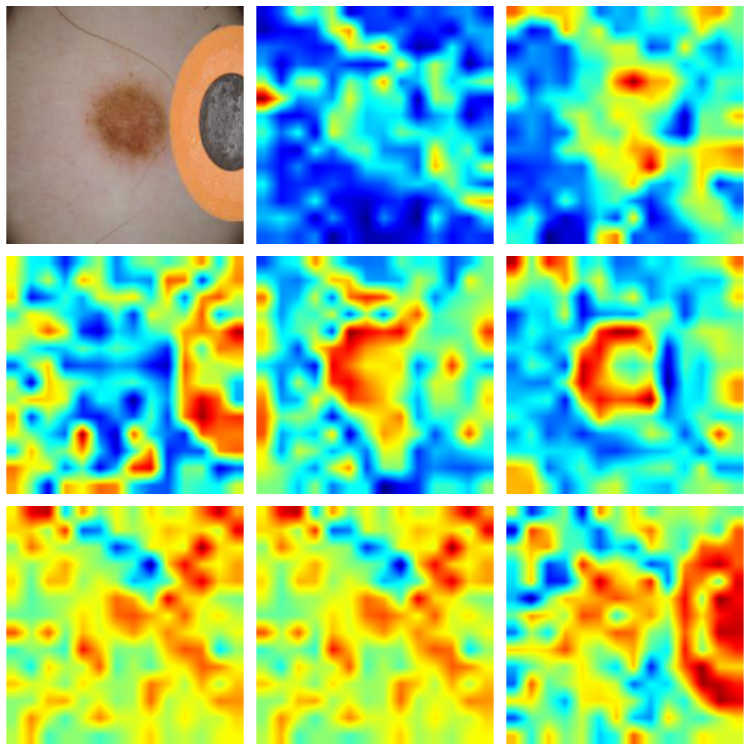}
\includegraphics[trim={0 0cm 0 0cm},clip,width=0.5\textwidth]{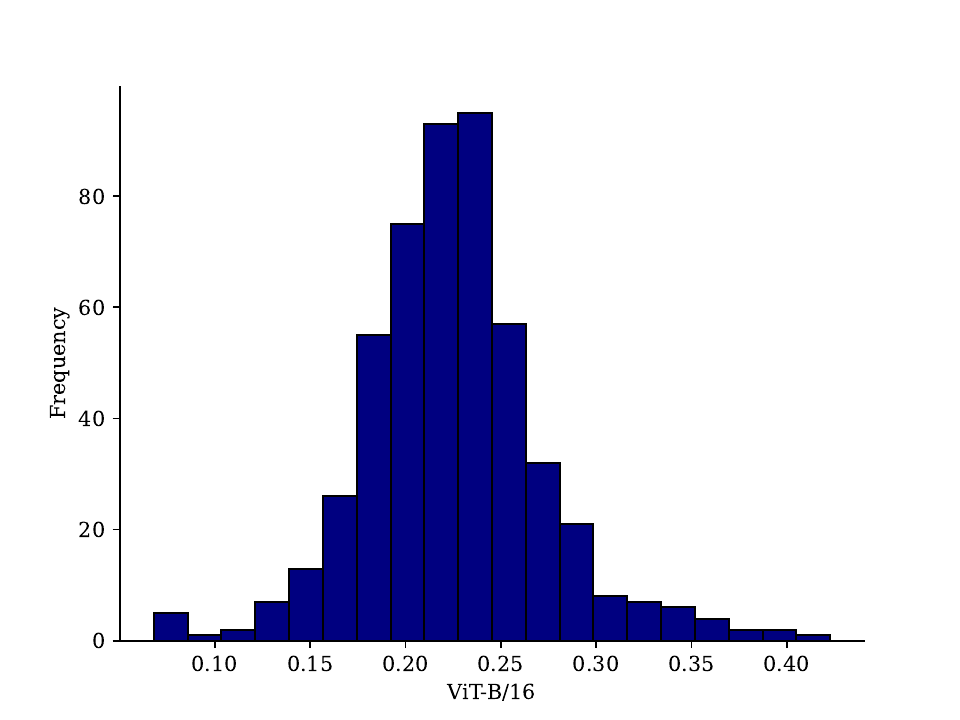}
\centering
\caption{Examples heatmap of neurons in the penultimate layer of \vit (left) and \sscore distribution (right).}
\label{fig:appendix-sscore-vit}
\end{figure}

In this section, we present the results of the analysis of whether there are spurious-encoding neurons in the representation space of ViTs. Figure~\ref{fig:appendix-sscore-vit} shows some examples of neuron heatmaps visualized with GradCam~\cite{gradcam:2020:ComVis} (left) and the distribution of \sscore averaged over 50 training inputs (right). We observe that even though there are some cases where the neuron focuses more on the core or spurious feature (more red region), the focus region of neurons in ViTs tends to be more distributed when projecting to the input. This leads to low overall \sscore, i.e., 0.23 for \isic and 0.14 for \wb (cf. Table~\ref{tab:appendix-dfr-vit}) - which are both in low-range \sscore. We hypothesize that this phenomenon is due to the ability of global learning from the multi-heads self-attention of ViTs. This mechanism allows the model to simultaneously attend to both core and spurious features across the entire input, distributing the learned representations. In Section~\ref{ssec:vit:attention-head}, we present an example in \isic that under the influence of highly spurious correlations, ViTs show a clear entangled relationship between core and spurious objects in some particular attention head.

On the other hand, in alignment with the conclusion for mitigation methods that leave the latent space unchanged in CNNs (cf. Section~\ref{ssec:cnn:entanglement}), DFR~\cite{Kirichenko:2023:ICLR} works well with \vit (cf. Table~\ref{tab:appendix-dfr-vit}). While significantly improving WGA in both \isic and \wb, the average \sscore of models before and after applying DFR are nearly identical. 
This again confirms that DFR indeed changes the way neurons interact with each other to improve the performance of a particular group rather than truly eliminating the learned spurious features.

\begin{table}[tbh]
    \centering
    \caption{Application DFR on \vit}
    \label{tab:appendix-dfr-vit}
    \begin{tabular}{p{0.18\textwidth}p{0.15\textwidth}p{0.15\textwidth}p{0.15\textwidth}p{0.15\textwidth}}
    \toprule
        & Model & WGA & Prune$_{h}$ & Avg. $\mathbf{s}$-score\\
         \midrule
         \multirow{2}{*}{ISIC} & Baseline & 0.16 & -  & 0.23\\
         & DFR & 0.76 & 89\% &   0.26\\
       \midrule
         \multirow{2}{*}{\wb} & Baseline & 0.66 & - &  0.14\\
        & DFR & 0.86 & 37\% & 0.14\\
        \bottomrule
    \end{tabular}
\end{table}

\paragraph{Takeaways.} With ViTs, we can not find a similar phenomenon of clear spurious-encoding neurons in the last layer as in CNNs. However, adapting neuron interactions works, suggesting the group-beneficial patterns when re-combining neurons in its representation.

%% file: 06-2_vit-attention-heads.tex
\subsection{Spurious Features in Attention Heads} \label{ssec:vit:attention-head}
The multi-head attention mechanisms in ViTs are designed to capture global information more effectively. We, therefore, investigate whether spurious and core features are disentangled in ViTs' attention heads.

\input{05_fig_attentionscores}

We analyze how much influence each image patch has on the encoding of one specific image patch, which either encodes a spurious (Figure~\ref{fig:attention-scores}a) or a core feature (Figure~\ref{fig:attention-scores}b). 
To visualize the attention map for a specific patch, we compute the attention weights in the transformer layers (averaged over multi-attention heads) when forwarding the input image through the model. From the attention weight matrix, we take the target row (indicating the target input patch) and visualize it as a heatmap.
Figure~\ref{fig:attention-scores} shows the visualizations across all transformer layers and single heads for a core feature in layer 9.
We observe that even though the target patch is a core feature (see Figure~\ref{fig:attention-scores} a), some specific layers of \vit show high attention to the spurious region (e.g. layer 8, layer 9) and that some neurons focus on the spurious region (e.g., attention heads 4, 5, 6, 12).

\paragraph{Takeaways.} Under the influence of spurious correlations, we observe that ViT jointly encodes information of core and spurious features.

%% file: 05_fig_attentionscores.tex
\begin{figure}[tbh]
    \centering
    \includegraphics[width=\linewidth]{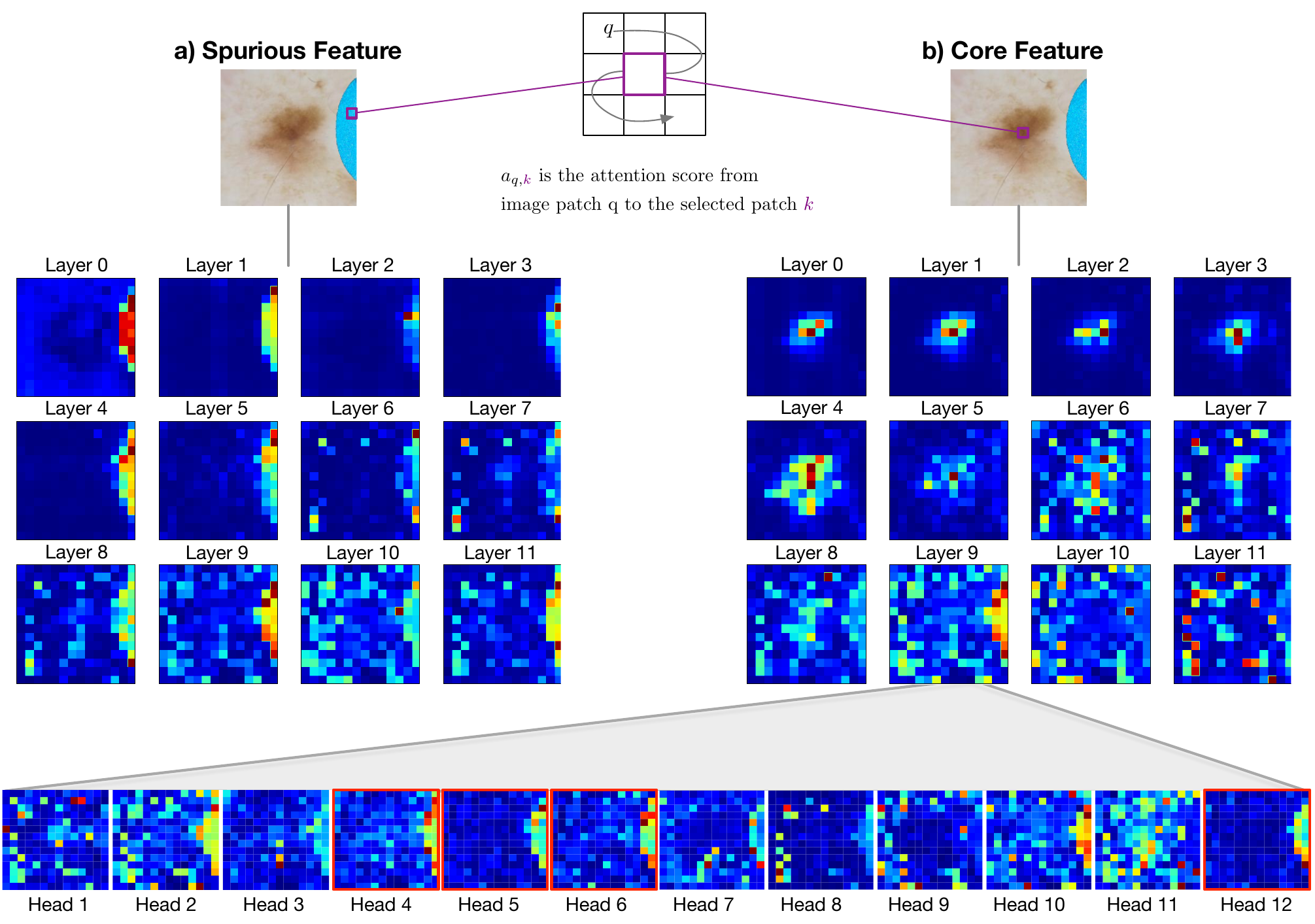}
    \caption{Attention scores in \vit of image patches corresponding to a) a spurious input feature and b) a core input feature.
    The feature maps in the middle show the scores per layer averaged over all attention heads.  For the spurious patch (b), the highest average attention scores are from other spurious patches, making the outline of the spurious input part visible in the attention map. For an image patch from a lesion (a), the highest average attention scores are from image patches representing skin or lesions. However, there are layers that encode both core and spurious features because some attention heads are focused on the spurious feature (e.g., attention heads 4,5,6, and 12 in layer 9, highlighted with a red border).}
    \label{fig:attention-scores}
\end{figure}

%% file: 07_discussion.tex
\section{Discussion} \label{sec:discussion}
This section summarizes and discusses our findings from the experiments in Sections~\ref{sec:sc-learning} to \ref{sec:vit}; a concise overview is given in Table~\ref{tab:cnn_vit_comparison}.

\paragraph{Vision models exploit spurious correlations.} We find that both CNNs and ViTs are susceptible to learning spurious features, and the ratio of minority and majority groups also affects the worst group accuracy. We also found that the spurious features can strongly define the representation space.

\paragraph{To some extent, there are disentangled spurious-encoding components.} In the representation layer, \sscore can be used to determine whether neurons in the penultimate layer of the models can be separated into highly spurious-encoding neurons. Considering the deeper layer of the models, previous work~\cite{csordas:2020:ICLR} proves that within a trained model there are sets of neurons that are solely responsible for a specific class of data. Using a similar technique, our result shows that it is possible to extract components (in CNNs) that are responsible for specific groups or features. Therefore, with a careful experimental design, we can find a component that, when removed, mainly eliminates the effect of spurious features. In ViTs, due to the lack of an equivalent technique, we cannot further analyze whether there really exist spurious-encoding components. The existence of disentangled spurious-encoding components leads to a simple pruning approach to eliminate the learned spurious correlations.

\paragraph{Without removing spurious-encoding neurons, adjusting the interaction between neurons can effectively improve performance.} We found that the effectiveness of last-layer re-weighting methods does not come from really eliminating the learned spurious features, but from reconstructing the interaction between neurons. These methods may remove unnecessary neurons, but not necessarily spurious-encoding neurons. This leads to the assumption that the way neurons are combined during learning and predicting is important for generalization ability. However, we do not yet fully understand the patterns when combining neurons.

\paragraph{There are also components that encode multiple features and cause models to mis-learn.} In the representation space, there is a large proportion of neurons with mid-range \sscore. While this may be due to the limitations of the XAI-based technique (cf. Appendix~\ref{app:sscore}), previous work~\cite{Le:2023:llr} has presented a qualitative example. In deeper components within CNNs, we find that removing spurious-encoding components also decreases the performance of the uncorrelated class. In multi-headed self-attention mechanisms, we find explicit cases where models fail to distinguish between core and spurious features. This suggests that in addition to a disentangled set of neurons encoding only core or spurious features, there are still neurons that activate both patterns.

\input{Tables/07_cnn-vit-comparison}

\paragraph{Limitations of existing methods for mitigating spurious feature learning.}
Considering existing post-hoc spurious reduction methods, we hypothesize that: (i) last-layer re-weighting methods could prune a large fraction of neurons, but in fact still need information from all types of encoded features, such as core and spurious. Therefore, these methods would fail in the severe case that models under-learn information, which is hard to detect when the overall performance is still high based on learning spurious features. Furthermore, unmodified representation learning is ultimately not optimal because these methods cannot \emph{correct} the earlier disentanglement of the model. (ii) For post-hoc pruning methods, we might prune some components or connections that are important for both core and spurious features. This could lead to the trade-off of improving WGA by reducing sensitivity to spurious features, but degrading performance for other groups by removing neurons that contribute to invariant features.

%% file: Tables/07_cnn-vit-comparison.tex
\setlength{\tabcolsep}{8pt}
\begin{table}[t!]
    \caption{Overview of shortcut learning in CNNs and ViT.}
    \label{tab:cnn_vit_comparison}
    \centering
    \renewcommand{\arraystretch}{1.5}
    \begin{tabular}{+p{3cm}^p{3.7cm}^p{3.7cm}}
        \toprule
        \tabhead  & \tabhead CNNs & \tabhead ViTs \\
        \otoprule
        Prevalence of shortcuts & \multicolumn{2}{p{7.4cm}}{Learn spurious correlations, shown by a large gap between worst-group and average accuracy.} \\
        \otoprule
        Shortcut features in latent space & Classes separate clearly, but minority groups are clustered. & Latent space is more defined by spurious features than classes. \\
        \otoprule
        Last-layer spurious-encoding neurons & Neurons show a wide range of \sscore, making disentanglement easier. & Spurious-encoding neurons are harder to detect, with average low \sscore. \\
        \otoprule
        Spurious neuron combinations in last layer & \multicolumn{2}{p{7.4cm}}{Adapt neurons interaction, not necessarily remove spurious-encoding neurons, can form a more robust combination against spurious correlation.} \\
        \otoprule
        Disentanglement in earlier components & Exist components encoding both spurious feature and uncorrelated class. & Exist attention heads show high attention score between core and spurious patch regions. \\
        \bottomrule
    \end{tabular}
\end{table}

%% file: 08_related-work.tex
\section{Related Work}
\label{sec:related-work}
A large body of work is based on the hypothesis that machine learning models tend to learn spurious features while under-learning invariant features. Therefore, a straightforward direction to mitigate the spurious correlations in learning is to focus on training group-robust models. Multiple approaches rely on human-annotated group labels and train models to minimize group loss~\cite{Arjovsky:2019:irm,Bao:ICML:2021,Sagawa:2020a:ICLR}. To reduce the cost of collecting human-annotated group labels, many studies propose estimating pseudo-group information using predictions from early-stopped ERM models, followed by training a second robust model with pseudo-group labels~\cite{Creager:2021:ICML,Liu:2021:ICML,Nam:2020:NeurIPS,Nam:2022:ICLR,ZhangMichael:2022:ICML}. Those methods assume that samples mis-classified by early-stopped ERM models are not holding spurious features. While these methods effectively improve the group-specific accuracy, they require expensive retraining, particularly when the existence of spurious information is unknown beforehand. 

Recent studies have shown that despite the strong correlation between spurious features and target labels, machine learning models can successfully learn high-quality spurious and core features~\cite{Kirichenko:2023:ICLR}. Therefore, post-hoc spurious mitigation or eliminating spurious correlations of trained models without extensive feature learning can be sufficient. Leaving the learned representation unchanged, existing work re-weights the classifier weights based on a group-balanced held-out dataset~\cite{Kirichenko:2023:ICLR} or searches for and only adjusts a single weight that affects the minority group most~\cite{Hakemi:2025}. Alternatively, it is possible to extract a subnetwork from a trained model that is more robust to spurious correlations~\cite{Le:2024:spuriousity,Park:2023:CVPR}. These works empirically demonstrate the potential of models to learn both core and spurious features. However, it remains unclear whether these features are truly disentangled and to what extent the underlying assumptions hold.

%% file: 09_conclusion.tex
\section{Conclusion} \label{sec:conclusion}
In this paper, we showed that both CNNs and ViTs are susceptible to spurious correlations. We provide evidence that models can learn to disentangle spurious features, allowing us to extract neurons or subnetworks within a trained network that are specifically responsible for these features. However, models may also encode a mixture of core and spurious features or fail to learn perfect disentanglement. Based on the results, this paper shows how and why methods mitigating spurious features that either leave the representation unchanged or extract only subnetworks from frozen trained weights may fail.

%% file: appendix.tex
\appendix
\section{Limitation of \sscore} \label{app:sscore}
\begin{wrapfigure}{l}{0.5\textwidth}
\includegraphics[trim={0 0cm 0 0cm},clip,width=0.4\textwidth]{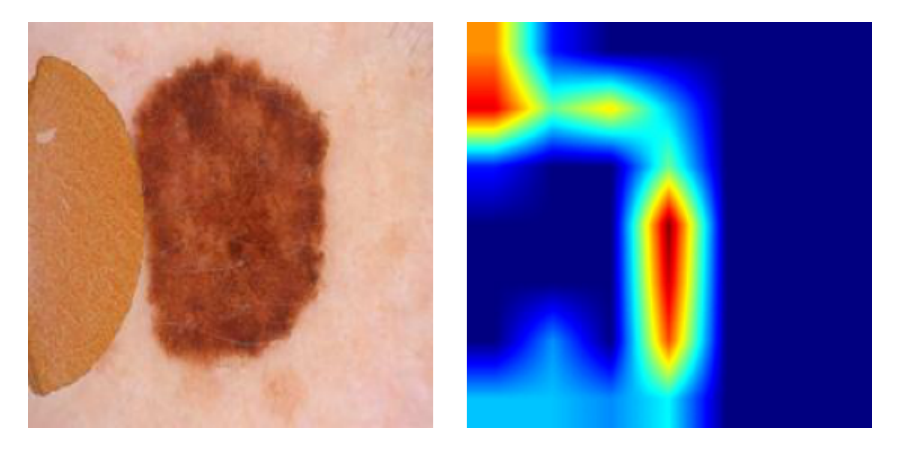}
\centering
\caption{.}
\label{fig:appendix_heatmap}
\end{wrapfigure}
In this section, we discuss the limitations of using post-hoc explainable AI methods to calculate \sscore. Explanation methods sometimes fail to capture fully what models are truly encoding. While heatmap attributions can visualize where the model is focusing within the input, they do not reveal what features the model is learning. For example, in Figure~\ref{fig:appendix_heatmap}, the \sscore is low because the neuron’s focus region has little overlap with the patch itself. However, to a human observer, it seems that the neuron is actually focusing on the \emph{edge of the patch}.